\documentclass[runningheads]{llncs}
\usepackage[T1]{fontenc}
\usepackage{graphicx}
\usepackage{float} 
\usepackage{subfigure}
\usepackage{amsmath}
\usepackage{amssymb}
\usepackage{marvosym}
\usepackage{hyperref}
\hypersetup{
    colorlinks = true,
    linkcolor=blue,
    filecolor=blue,      
    urlcolor=blue,
    citecolor=blue,
}
\begin{document}
\title{ Toward Clinically Assisted Colorectal Polyp Recognition via Structured Cross-modal Representation Consistency  }
\titlerunning{Toward CPC via Structured Cross-modal Representation Consistency}
\author{Weijie Ma\inst{1} \and Ye Zhu\inst{1} \and Ruimao Zhang\inst{1} 
\thanks{Corresponding Author}
\and Jie Yang\inst{1} \and Yiwen Hu \inst{1} 
\and Zhen Li\inst{1} \and Li Xiang\inst{2}
}
\authorrunning{W. Ma et al.}
\institute{Shenzhen Research Institute of Big Data, The Chinese University of Hong Kong (Shenzhen), China \\
\email{\{weijiema@link, zhangruimao@\}cuhk.edu.cn}
\and Longgang District People's Hospital of Shenzhen, China}
\maketitle              % typeset the header of the contribution
\begin{abstract}
The colorectal polyps classification is a critical clinical examination. To improve the classification accuracy, most computer-aided diagnosis algorithms recognize colorectal polyps by adopting Narrow-Band Imaging (NBI). However, the NBI usually suffers from missing utilization in real clinic scenarios since the acquisition of this specific image requires manual switching of the light mode when polyps have been detected by using White-Light (WL) images.
To avoid the above situation, we propose a novel method to directly achieve accurate white-light colonoscopy image classification by conducting structured cross-modal representation consistency. In practice, a pair of multi-modal images, \textit{i.e.} NBI and WL, are fed into a shared Transformer to extract hierarchical feature representations. Then a novel designed Spatial Attention Module (SAM) is adopted to calculate the similarities between class token and patch tokens %from multi-levels 
for a specific modality image. By aligning the class tokens and spatial attention maps of paired NBI and WL images at different levels, the Transformer achieves the ability to keep both global and local representation consistency for the above two modalities. Extensive experimental results illustrate the proposed method outperforms the recent studies with a margin, realizing multi-modal prediction with a single Transformer while greatly improving the classification accuracy when only with WL images. Code is available at \href{https://github.com/WeijieMax/CPC-Trans}{\text{https://github.com/WeijieMax/CPC-Trans}}.

\end{abstract}

\section{Introduction}
Adenomatous polyp is considered to be the underlying cause of colorectal cancer (CRC) \cite{ref2}, which is the second lethal cancer and the third most commonly diagnosed malignancy \cite{ref3}. Detection and resection of the polyps usually depend on colonoscopy. 
However, in clinical practice, the standard white-light (WL) observation could only provide limited discriminative information between neoplasticism and nonneoplastic colorectal polyps.

To improve classification accuracy, computer-aided diagnosis systems (CADx) are introduced, most of which adopt Narrow-Band Imaging (NBI), Blue Light Imaging (BLI) and other enhanced colonoscopy images \cite{ref6,ref5}. For example, Fonolia et al. \cite{ref7} proposed a CADx system for classifying colorectal polyps combining WL, BLI and Linked Colour Imaging (LCI) modalities, which achieved encouraging performance. Moreover, Franklin et al. \cite{ref8} introduced a robust frame-level strategy using NBI sequences to learn a deep convolutional representation, which achieved an average classification accuracy of 90.79\%.

Despite the enhanced imaging, endoscopists often rely on WL images before they change the light mode to detect the possible polyps, which means that the WL images may fail to capture discriminative features of polyps, resulting in much less accurate classification by other imaging means. Therefore, a well-designed WL-based CADx system is in urgent need for better colorectal polyp recognition using only WL images. In recent studies, a deep learning model proposed by Yang et al. \cite{ref9} presented a promising performance in classifying colorectal lesions only on WL colonoscopy images. To further narrow the gap between conventional WL images and enhanced images, Wang et al. \cite{ref10} enhanced low representation WL features with NBI images through domain alignment and contrastive learning which improved the WL-based classification results.

In this paper, we propose a simple yet effective method to improve the discriminative representation of WL images by conducting structured cross-modal representation consistency, realizing accurate WL colonoscopy image classification. 
During the training, the hierarchical feature representations of the paired WL and NBI images are extracted by a shared Transformer.
A carefully designed novel Spatial Attention Modules (SAM) is further exploited to align the output class tokens of the paired images, while their spatial attention maps are further constrained to be consistent at different levels,
enhancing the feature representation of WL images from the global and local perspectives simultaneously.

The main contributions are three-fold. (1) An effective scheme based on global and local consistent learning is proposed to extract more discriminative representations of WL images for colorectal polyps classification (CPC). 
(2) We introduce a general framework of multi-modal learning for medical image analysis based on Transformer where the unilateral advantages can be propagated across domains through introducing external attention modules but dropped out these external auxiliary modules in inference, efficiently reducing the model complexity in applications. 
(3) The proposed method outperforms state-of-the-arts of CPC by a margin, promoting the development of related methods in clinically assisted colorectal polyp recognition.

\section{Related Work}
\subsubsection{Domain Alignment.} Domain Alignment mainly focuses on reducing the discrepancy between the associated distributions of the projected source and target features. Two different schemes are usually considered for aligning feature representations based on recent deep domain alignment methods: (i) extracted features are learned in a share subspace, aiming to minimize the distance between the source and the target distributions. 
(ii) Another scheme concentrates on reducing the Maximum Mean Discrepancy and is commonly adopted in the case of missing target domain labels. In \cite{ref12}, a nonlinear transformation is learned to align correlations of layer activations in deep neural networks (Deep CORAL). Moreover, by utilizing the intra-class variation in the target domain, Chen et al. \cite{ref13} proposed the Progressive Feature Alignment Network to align the discriminative feature across different domains. 

\subsubsection{Attention and Vision Transformer.} 
To gain a more vivid representation of related different positions from a single sequence or sentence, the attention mechanism was redefined as the self-attention and later adopted by the Transformer architecture \cite{ref16} that based solely on it. Soon after that, extensive works \cite{dvct,sanet,scan} have been improved by leveraging the attention mechanism. Currently, Dosovitskiy et al. \cite{ref17} applied Transformer architecture from NLP to computer vision as Vision Transformer (ViT), and showed that the sequences of image patches could perform very well with a pure transformer on image classification, while the convolutional networks usually suffer from difficulty in capturing and storing long-distance dependent information due to the limited receptive field.  %Following the ViT, the data-efficient image transformer (DeiT) \cite{ref18} was proposed to help training the Vision Transformer with a novel distillation procedure. 
Following the ViT, many other vision transformer variants are proposed \cite{ref19,ref20}, and some of them have achieved great performance on various medical tasks \cite{swinunet,mctrans,cotr,yangjie,nnformer} with the strong representation capabilities of transformer.

\section{Method}
\subsection{Problem Formulation and Framework Overview}
Given a WL image and its corresponding NBI image \{$I_{w}$, $I_{n}$\} with their shared label $G$, 
the proposed cross-modal learning framework aims to explore structured semantic information by utilizing a shared class proxy embedding that bridges the modality gap after the training process. Finally, this framework produces a unified accurate CPC model for WL-only predictions, outperforming ones trained with single-modal data with a margin.

\begin{figure}[H]
\centering 
\includegraphics[width=1\textwidth]{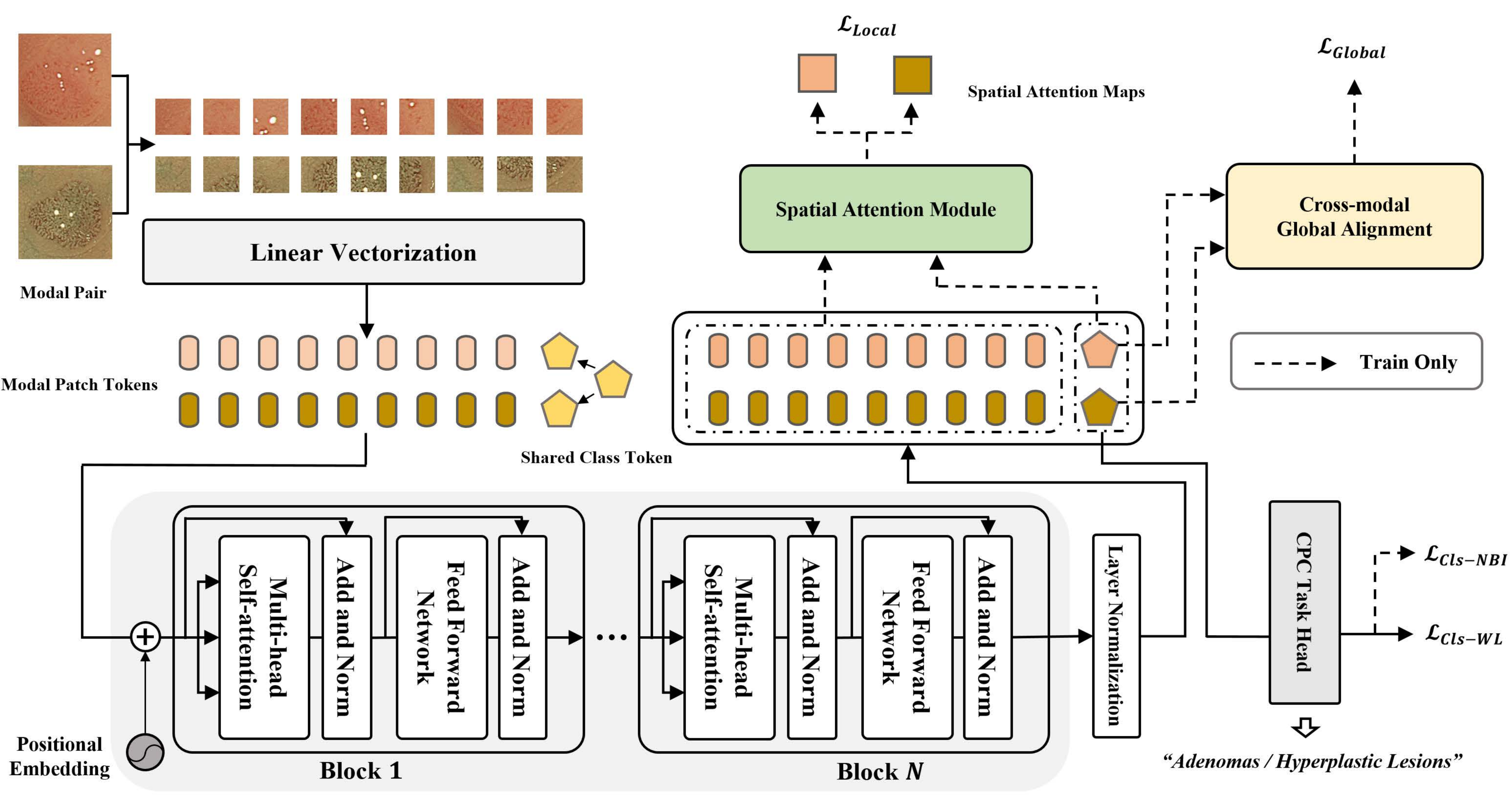} 
\caption{Overview of our proposed cross-modal shared colorectal polyp recognition framework via single Transformer
architecture and the proposed CGA and SAM. Note that all parts linked by dotted lines can be removed during the inference phase.} 
\label{Fig.main2} 
\end{figure}

As illustrated in Fig.~\ref{Fig.main2}, we introduce a Transformer-based framework with cross-modal global alignment (CGA) and spatial attention module (SAM). We first feed a dual-modal image pair as input $\{I_{w}, I_{n}\} \in \mathbb{R} ^ {H \times W \times 3} $ to the framework. The image-pair is then divided into $P \times P$ image patches. After a linear projection layer, each patch is embedded into a patch token with embedded dimension $d = \frac{3P^2}{2}$. Note that here we reduce the embedded dimension to $1/2$ as standard so as to release the computational overhead and drop model parameters. Patch embedding is implemented by a convolutional layer of a $P \times P$ kernel. In ~this ~way, ~we could reassign the dimension of the ~patch ~token ~inputs ~as

\noindent  $\{X_{w}, X_{n}\} \in \mathbb{R} ^ {N \times d}$, where $ N = \frac{HW}{P^2}$. Then, $\{X_{w}, X_{n}\}$ will be successively concatenated with a shared learnable class token $\textbf{c} \in \mathbb{R}^{d}$. To compensate for the missing 2-D structural information, positional embedding $E \in \mathbb{R}^{(N+1) \times d}$ is supplemented to $\{X_{w}, X_{n}\}$ and $\textbf{c}$ by element-wise addition. As a result, information about the relative or absolute position is provided in patch tokens. 

Next, dual-modal patch tokens are separately passed through a series of shared Transformer blocks for deep feature extraction and comprehensive context modeling to generate modality-specific global representations on two class tokens $\{ \textbf{c}_{w}, \textbf{c}_{n} \} \in \mathbb{R} ^ {d}$. During training, $\{ \textbf{c}_{w}, \textbf{c}_{n} \}$ are then fed into CGA to align dual-modal image pair's global representation. Meanwhile, SAM furthers cross-modal local alignment by comparing two modalities' response maps between their global representation and local instance-level information. 

The inference stage only relies on the single Transformer architecture and discards any modality-specific or cross-modal modules for CPC prediction.

\subsection{Shared Transformer Block}
The shared Transformer blocks are employed to learn pixel-level contextual dependencies adapted to both modalities. As shown in Fig.~\ref{Fig.main2}, the share Transformer blocks consist of $B$ layers, and each is composed of two components, multi-head self-attention (MSA) and a feed-forward network (FFN). Layer normalization (LN) is applied before each component while skipping connection after each component. 

The FFN is a multi-layer perceptron (MLP) which includes two linear layers with a medium GeLU activation. In addition, considering a pair of arbitrary dual-modal token sequences $\{X_{w}, X_{n} \} \in \mathbb{R}^{\hat{N} \times d}$ as input, single-head self-attention (SA) formulated as:

\begin{equation}
    {\rm SA}(X^{i}_{m}) = {\rm Softmax}( \frac{X^{i}_{m}\textbf{W}^i_Q(X^{i}_{m}\textbf{W}^i_K)^{\rm T}}{\sqrt{d^{\prime}}})(X^{i}_{m}\textbf{W}^i_V) 
\label{eqn:1}
\end{equation}

\noindent where $m \in \{w, n\}$. $\{\textbf{W}^i_Q, \textbf{W}^i_K, \textbf{W}^i_V \} \in \mathbb{R}^{d \times d^{\prime}}$ donate the triple parameter matrices of $i^{th}$ layer and $d^{\prime}$ is the dimension of each head. Here, we omit LN layers for simplicity. MSA is a concatenation of $h$ parallel SA modules with linear projection to rearrange their outputs. In our experiments, we employ $h = 6$, $d = 384$, and $d^{\prime} = d/h = 64$. As depicted in Fig.~\ref{Fig.main2}, the whole calculation can be formulated as:

\begin{equation}
    X^{i}_{m} = {\rm MSA}(X^{i-1}_{m}) + {\rm FFN}(X^{i-1}_{m} + {\rm MSA}(X^{i-1}_{m}))
\label{eqn:2}
\end{equation}

\subsection{Cross-modal Global Alignment}

Diminishing the discrepancy between WL and NBI images allows a model to leverage cross-modal knowledge from either of modalities. 
To this end, we propose Cross-modal Global Alignment (CGA), an auxiliary module only used during the training stage. 
CGA maintains the cross-modal consistency in the shared model, or in other words, drives the model to learn cross-domain knowledge while only WL images are given. 

In practice, to begin with, CGA computes the cosine difference between two modal-specific class tokens of two paired images. Then, imposing a loss function on their cosine similarity reduces the average cosine distance between image-pairs of two modalities. This way, feature representations from paired images may match up more closely, and the model may learn to capture hard features from NBI images (\textit{e.g.} textures clear in NBI images but unclear in WL images).

As figured in Fig.~\ref{Fig.main2}, after being passed through an external LN, the average distance between two modalities' paired samples is narrowed based on cosine similarity. Given two modality-specific class tokens $\{ \textbf{c}_{w}, \textbf{c}_{n} \}$, their alignment loss is computed by:

\begin{equation}
    \mathcal{L}_{global} = 1 - \rm{dist}_{cos}(\textbf{c}_w, \textbf{c}_n) = 1- \frac{\textbf{c}_w \cdot \textbf{c}_n}{|| \textbf{c}_w ||_2 || \textbf{c}_n ||_2 }
    \label{eqn:3}
\end{equation}

\subsection{Spatial Attention Module}
Despite the global alignment of modal-specific class tokens, we also propose the Spatial Attention Module (SAM) to pursue the multi-level structured semantic consistency between two modalities. First, we obtain through SAM the globally guided affinity, \textit{i.e.}, the response map between each image's global representation and local regions. Subsequently, we align two modalities' local semantics by limiting the distance between two modalities' response maps.

Concretely, as illustrated in Fig.~\ref{Fig.main2}, SAM takes as input the output patch-token features from the external ${\rm LN}$ layer \{$F_w, F_n$\} $\in \mathbb{R}^{N \times d}$ as well as the modal-specific class tokens \{$\textbf{c}_w$, $\textbf{c}_n$\} $\in \mathbb{R}^d$. In practice, we utilize \{$F_w$, $F_n$\} to generate the key of Spatial Attention (SpA) operation through linear projection, while the query of SpA is obtained based on \{$\textbf{c}_w$, $\textbf{c}_n$\}, described as follows:

\begin{equation}
    R_m = \rm{SpA}(F_m, \textbf{c}_m) = \rm{Softmax}(\textbf{c}_\textit{m}\textbf{W}_\textit{Q} (F_m\textbf{W}_\textit{K})^T)
    \label{eqn:4}
\end{equation}

\noindent where $m \in \{w, n\}$. \{$R_w$, $R_n$\} $\in \mathbb{R}^N$ are two response maps for WL and NBI images respectively, representing global-to-local affinity of each modality. Similar to Eqn.~\ref{eqn:3}, we compute cosine similarity between two response maps by: 

\begin{equation}
    \mathcal{L}_{local} = \lambda (1-{\rm{dist}}_{\rm{cos}}(R_w, R_n))
    \label{eqn:5}
\end{equation}

\noindent where $\lambda$ is a ratio coefficient to normalize the magnitude of $\mathcal{L}_{local}$ to balance it with $\mathcal{L}_{global}$ (here we set $\lambda$ = 0.3). Taking the global representation as standard, the distribution of semantics over local regions is obtained by Eqn.\ref{eqn:4}. In Eqn.\ref{eqn:5}, $\mathcal{L}_{local}$ computes the distance between two global-to-local affinity distributions. Here, as two modalities' local semantics are extracted by a shared model, they are in shared representation space. Therefore, by minimizing $\mathcal{L}_{local}$, SAM could align two modalities in terms of their local semantics. Further, with local semantics bridged between WL and NBI image pairs, the model is learning to mine hidden local features from WL images, thus bridging the accuracy gap between them. Arguably, in our experiments, SAM further brings a 0.7\% accuracy improvement. %

In this way, the overall training loss contains four parts:

\begin{equation}
    \mathcal{L} = \mathcal{L}_{cls-WL} + \mathcal{L}_{cls-NBI} + \mathcal{L}_{Global} + \mathcal{L}_{Local}
    \label{eqn:6}
\end{equation}

\noindent in which $\mathcal{L}_{cls-WL} = CrossEntropy(H(\textbf{c}_w)),$ $ \mathcal{L}_{cls-NBI} = CrossEntropy(H(\textbf{c}_n))$ to supervise the CPC binary classification task ($i.e.,$ adenomatous or hyperplastic) given the ground truth label $G$. Note that at the inference stage, the model performs binary classification for WL images only.

\section{Experiments and Results}

\subsection{Dataset}
The dataset we used is named CPC-Paired Dataset \cite{ref10}. The dataset has two paired image modalities (WL-NBI) and consists of two parts due to the insufficient amount of each part. Every WL image and corresponding NBI image share the same label of a colorectal polyp. There are 307 image pairs labeled as adenomas and 116 images labeled as hyperplastic lesions in total. One part of the dataset was extracted from ISIT-UMR Colonoscopy Dataset \cite{ref22} and another part was collected from the hospital. Specifically, the ISIT-UMR part contains 102 adenomatous and 63 hyperplastic paired frames of two modals, collated by 40 adenomas and hyperplastic lesions sequences from a total of 76 short categorized video data. And the clinical part was composed of 258 WL-NBI pairs, 205 adenoma images, and 53 hyperplastic polyp images in detail from 123 patients. What's more, the dataset has annotated bounding box data for subsequently cropping lesion area corresponding to all images of two modalities.

\begin{table}[!ht]
    \centering
    \caption{Comparisons of our full model with previous best studies on the test sets with respect to accuracy metric. $\ddag$~refers to that the result is directly cited from the original paper based on their own divisions of training-validation sets. }
    
    \label{table:1}
    \begin{tabular}{lcccccccc}
    \hline
        Methods & Backbone & Params (M) & Fold 1 & Fold 2 & Fold 3 & Fold 4 & Fold 5 & Mean \\ \hline
        ~ & VGG & 138.36 & 78.2\% & 79.5\% & 77.3\% & 78.0\% & 77.9\% & 78.2\% \\ 
        Yang \cite{ref9}$^\ddag$ & InceptionV3 & 24.73 & 81.1\% & 82.1\% & 80.5\% & 82.0\% & 81.3\% & 81.5\% \\ 
        ~ & ResNet50 & 22.56 & 79.5\% & 80.5\% & 78.0\% & 80.3\% & 78.8\% & 79.7\% \\ 
        Wang \cite{ref10}$^\ddag$ & ResNet50 & 22.56 & 85.9\% & 86.1\% & 84.2\% & 85.8\% & 85.2\% & 85.3\% \\ \hline
        Ours & ViT-Small & \textbf{21.67} & 87.2\% & 88.5\% & 94.3\% & 92.7\% & 75.8\% & \textbf{87.7}\% \\ \hline
        
    \end{tabular}
\end{table}

\subsection{Implementation Details}
In our work, we conducted 5-fold  cross-validation for all experiments. The training and validation set is 8 : 2, and the partition in each fold was generated  randomly based on every subject. We implemented our model with the PyTorch toolkit on a single NVIDIA V100 GPU. The input size of the image is 224 $\times$ 224, and the data augmentation was adopted by random resized cropping and horizontal flipping. Our model's embedding dimension is 384 with 6 attention heads and 12 block layers based on the setting of the ViT small version. We choose SGD optimizer and cosine annealing learning rate scheduler for network optimization, Following~\cite{ref10}, the batch size is 16, and learning rate is 1$e-$3 with a maximum of 500 epochs. We also use the momentum of 0.9 and the weight decay of 5$e-$5.

\subsection{Results and Analysis}
Through massive experiments, we verify the effectiveness of our proposed method. To be specific, as shown by a 5-fold cross-validation result in Table.~\ref{table:1}, despite the fewer parameters, we can observe that our approach outperforms existing state-of-the-art methods. This also proves the stronger representation power and higher parameter efficiency of our model.

The ablation study further examines the effectiveness of each component, which is shown in Table.~\ref{table:2}. ``Trans with WL-only'' means vanilla Transformer-

\begin{table}[H]
    \centering
    \caption{The performance of the baseline and our full model on the test sets with respect to accuracy metric. ``$\Delta$ M'' indicates the gain of current average result compared with the former row, and the first row is recent state-of-the-art result.}
    \label{table:2}
    \begin{tabular}{lcccccccc}
        \hline
        Methods & Fold 1 & Fold 2 & Fold 3 & Fold 4 & Fold 5 & Mean & $\Delta$ M \\ \hline
        Wang \cite{ref10} & 85.9\% & 86.1\% & 84.2\% & 85.8\% & 85.2\% & 85.3\% & $-$ \\ 
        Trans with WL-only & 84.6\% & 88.5\% & 93.1\% & 82.9\% & 76.5\% & 85.1\% & $-$ 0.2\% \\
        Trans + CGA & 87.2\% & 88.5\% & 93.1\% & 87.8\% & 77.4\% & 86.8\% & \textbf{+1.7\%} \\ 
        Trans + CGA + SAM & 87.2\% & 88.5\% & 94.3\% & 92.7\% & 75.8\% & \textbf{87.7\%} & \textbf{+0.9\%} \\ 
        \hline
    \end{tabular}
\end{table}

\begin{figure}[H] 

\centering 
\includegraphics[width=1\textwidth]{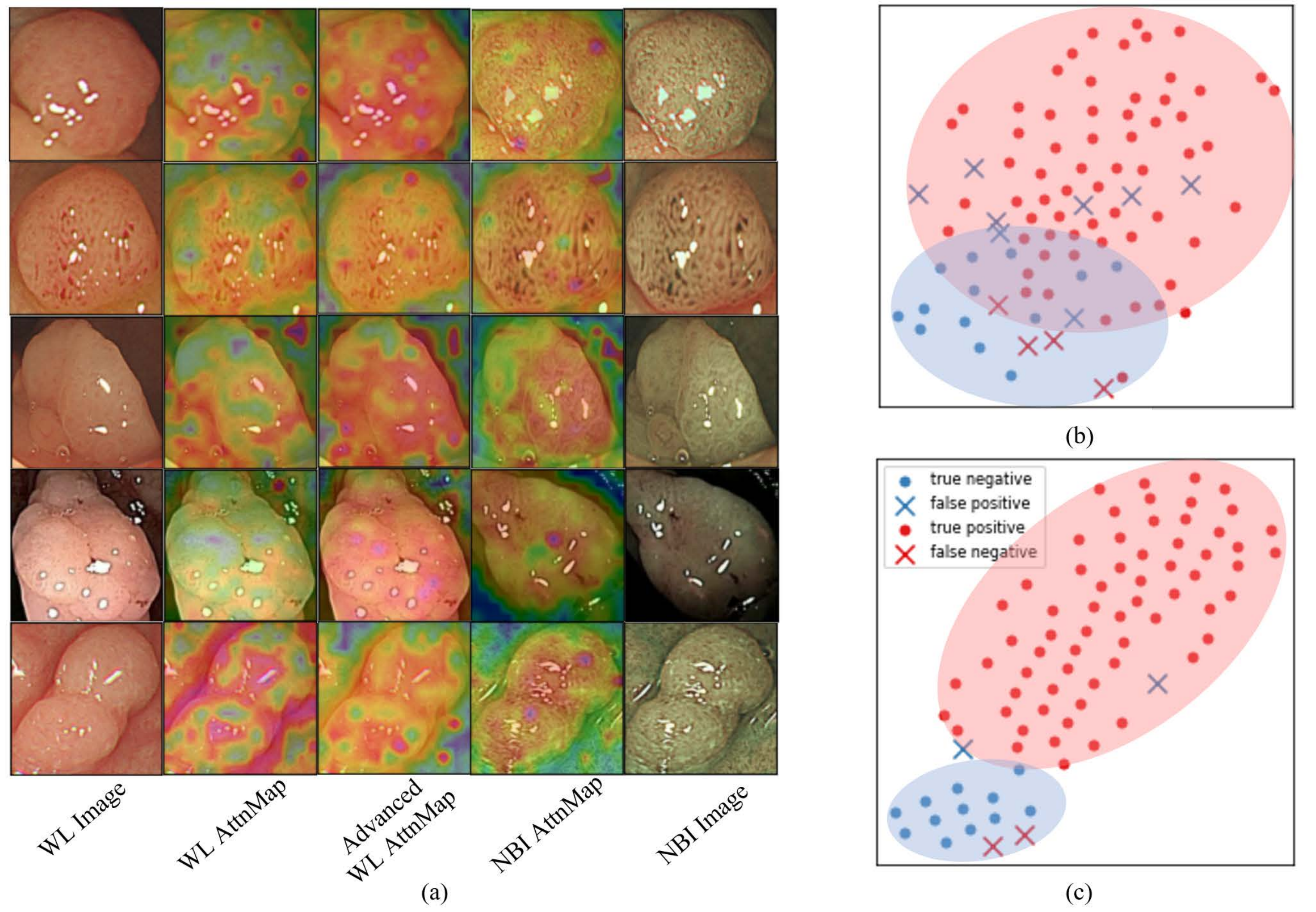} 
\caption{(a) is the comparative study of different attention maps. ``WL/NBI AttnMap'' is produced by the WL/NBI vanilla model and ``Advanced WL AttnMap'' by the proposed model. 
(b) displays the predictive distribution of WL images given by WL-only Transformer over the 2-dimensional representation space offered by t-SNE, while (c) shows the corresponding distribution obtained from our proposed shared Transformer. } 
\label{Fig.main3} 
\end{figure}

\noindent based baseline of WL images. ``Trans $+$ X'' indicates the proposed framework combined with only CGA or both CGA and SAM, which improve the baseline by 1.7\% or 2.6\%. This ablation study demonstrates that the accuracy gain is more heavily from our proposed modules other than the Transformer architecture.

The qualitative analysis is illustrated in Fig.~\ref{Fig.main3}. Precisely, we extract the attention maps of the last Transformer layer of our model and vanilla models for each modality and remap them on the original images~\cite{Gao_2021_ICCV}. As is shown in Fig.~\ref{Fig.main3} (a), the advanced WL attention maps from our model are more similar to the NBI attention maps, covering nearly the entire lesion region. In contrast, the WL vanilla one underperforms the former, focusing on fractional attentive areas, irrelevant corners, or backgrounds. There is an apparent visual gap between WL attention maps and our advances, proving our model's semantic consistency between the two modalities.

Fig~\ref{Fig.main3} (b)(c) respectively shows the WL-only model and our proposed model's discrimination between adenomatous or hyperplastic lesion through t-SNE visualization~\cite{van2008visualizing}. In Fig~\ref{Fig.main3} (b), there is an obvious overlapping region between two predictive distributions in which the WL-only model is dramatically less discriminative, resulting in a higher number of false samples (marked by forks). However, in Fig~\ref{Fig.main3} (c), our proposed model separates two distributions with a clear margin, with only two outliers for each class. This result shows that our proposed shared Transformer can achieve higher performance on white-light colonoscopy image classification by untangling two predictive distributions.  
\section{Conclusion}
In this paper, a novel Transformer-based framework is introduced to tackle WL-only CPC, which proposed the Cross-modal Global Alignment (CGA) and a newly designed Spatial Attention Module (SAM) to pursue the structured semantic consistency, \textit{i.e.} modality-aware global representations and instance-aware local correlations. In the inference phase, all of the modules used for modality alignment can be removed, guaranteeing WL-only simple but accurate prediction for CPC task. Extensive experimental results and visualizations demonstrate the effectiveness of our proposed multi-level consistency learning method.

\section{Acknowledgement}
The work is supported in part by the Young Scientists Fund of the National Natural Science Foundation of China under grant No. 62106154, by Natural Science Foundation of Guangdong Province, China (General Program) under grant No.2022A1515011524 and by the Guangdong Provincial Key Laboratory of Big Data Computing, The Chinese University of Hong Kong, Shenzhen.


\begin{thebibliography}{10}
    
    \bibitem{ref14}
    Bahdanau, D., Cho, K., Bengio, Y.: Neural machine translation by jointly
      learning to align and translate. arXiv preprint arXiv:1409.0473  (2014)
    
    \bibitem{ref6}
    Bisschops, R., Hassan, C., Bhandari, P., Coron, E., Neumann, H., Pech, O.,
      Correale, L., Repici, A.: Basic (bli adenoma serrated international
      classification) classification for colorectal polyp characterization with
      blue light imaging. Endoscopy  \textbf{50}(03),  211--220 (2018)
    
    \bibitem{swinunet}
    Cao, H., Wang, Y., Chen, J., Jiang, D., Zhang, X., Tian, Q., Wang, M.:
      Swin-unet: Unet-like pure transformer for medical image segmentation. arXiv
      preprint arXiv:2105.05537  (2021)
    
    \bibitem{ref19}
    Carion, N., Massa, F., Synnaeve, G., Usunier, N., Kirillov, A., Zagoruyko, S.:
      End-to-end object detection with transformers. In: European Conference on
      Computer Vision. pp. 213--229. Springer (2020)
    
    \bibitem{ref13}
    Chen, C., Xie, W., Huang, W., Rong, Y., Ding, X., Huang, Y., Xu, T., Huang, J.:
      Progressive feature alignment for unsupervised domain adaptation. In:
      Proceedings of the IEEE/CVF Conference on Computer Vision and Pattern
      Recognition. pp. 627--636 (2019)
    
    \bibitem{ref17}
    Dosovitskiy, A., Beyer, L., Kolesnikov, A., Weissenborn, D., Zhai, X.,
      Unterthiner, T., Dehghani, M., Minderer, M., Heigold, G., Gelly, S.,
      Uszkoreit, J., Houlsby, N.: An image is worth 16x16 words: Transformers for
      image recognition at scale. In: International Conference on Learning
      Representations (2021)
    
    \bibitem{ref7}
    Fonoll{\`a}, R., EW~van~der Zander, Q., Schreuder, R.M., Masclee, A.A., Schoon,
      E.J., van~der Sommen, F., et~al.: A cnn cadx system for multimodal
      classification of colorectal polyps combining wl, bli, and lci modalities.
      Applied Sciences  \textbf{10}(15), ~5040 (2020)
    
    \bibitem{Gao_2021_ICCV}
    Gao, W., Wan, F., Pan, X., Peng, Z., Tian, Q., Han, Z., Zhou, B., Ye, Q.:
      Ts-cam: Token semantic coupled attention map for weakly supervised object
      localization. In: Proceedings of the IEEE/CVF International Conference on
      Computer Vision. pp. 2886--2895 (2021)
    
    \bibitem{uxnet}
    Ji, Y., Zhang, R., Li, Z., Ren, J., Zhang, S., Luo, P.: Uxnet: Searching
      multi-level feature aggregation for 3d medical image segmentation. In:
      International Conference on Medical Image Computing and Computer Assisted
      Intervention. pp. 346--356 (2020)
    
    \bibitem{mctrans}
    Ji, Y., Zhang, R., Wang, H., Li, Z., Wu, L., Zhang, S., Luo, P.: Multi-compound
      transformer for accurate biomedical image segmentation. In: International
      Conference on Medical Image Computing and Computer Assisted Intervention. pp.
      326--336 (2021)
    
    \bibitem{ref3}
    Keum, N., Giovannucci, E.: Global burden of colorectal cancer: emerging trends,
      risk factors and prevention strategies. Nature reviews Gastroenterology \&
      hepatology  \textbf{16}(12),  713--732 (2019)
    
    \bibitem{ref5}
    Komeda, Y., Kashida, H., Sakurai, T., Asakuma, Y., Tribonias, G., Nagai, T.,
      Kono, M., Minaga, K., Takenaka, M., Arizumi, T., et~al.: Magnifying narrow
      band imaging (nbi) for the diagnosis of localized colorectal lesions using
      the japan nbi expert team (jnet) classification. Oncology  \textbf{93}(Suppl.
      1),  49--54 (2017)
    
    \bibitem{van2008visualizing}
    Van~der Maaten, L., Hinton, G.: Visualizing data using t-sne. Journal of
      Machine Learning Research  \textbf{9}(11) (2008)
    
    \bibitem{ref22}
    Mesejo, P., Pizarro, D., Abergel, A., Rouquette, O., Beorchia, S., Poincloux,
      L., Bartoli, A.: Computer-aided classification of gastrointestinal lesions in
      regular colonoscopy. IEEE transactions on Medical Imaging  \textbf{35}(9),
      2051--2063 (2016)
    
    \bibitem{ref8}
    Sierra-Jerez, F., Mart{\'\i}nez, F.: A deep representation to fully
      characterize hyperplastic, adenoma, and serrated polyps on narrow band
      imaging sequences. Health and Technology pp. 1--13 (2022)
    
    \bibitem{ref12}
    Sun, B., Saenko, K.: Deep coral: Correlation alignment for deep domain
      adaptation. In: European Conference on Computer Vision. pp. 443--450.
      Springer (2016)
    
    \bibitem{ref20}
    Touvron, H., Cord, M., Douze, M., Massa, F., Sablayrolles, A., J{\'e}gou, H.:
      Training data-efficient image transformers \& distillation through attention.
      In: International Conference on Machine Learning. pp. 10347--10357. PMLR
      (2021)
    
    \bibitem{ref16}
    Vaswani, A., Shazeer, N., Parmar, N., Uszkoreit, J., Jones, L., Gomez, A.N.,
      Kaiser, {\L}., Polosukhin, I.: Attention is all you need. Advances in neural
      information processing systems  \textbf{30} (2017)
    
    \bibitem{ref10}
    Wang, Q., Che, H., Ding, W., Xiang, L., Li, G., Li, Z., Cui, S.: Colorectal
      polyp classification from white-light colonoscopy images via domain
      alignment. In: International Conference on Medical Image Computing and
      Computer-Assisted Intervention. pp. 24--32. Springer (2021)
    
    \bibitem{dvct}
    Wang, T., Zhang, R., Lu, Z., Zheng, F., Cheng, R., Luo, P.: End-to-end dense
      video captioning with parallel decoding. In: Proceedings of the IEEE/CVF
      International Conference on Computer Vision. pp. 6847--6857 (2021)
    
    \bibitem{ref21}
    Wang, W., Xie, E., Li, X., Fan, D.P., Song, K., Liang, D., Lu, T., Luo, P.,
      Shao, L.: Pyramid vision transformer: A versatile backbone for dense
      prediction without convolutions. In: Proceedings of the IEEE/CVF
      International Conference on Computer Vision. pp. 568--578 (2021)
    
    \bibitem{sanet}
    Wei, J., Hu, Y., Zhang, R., Li, Z., Zhou, S.K., Cui, S.: Shallow attention
      network for polyp segmentation. In: International Conference on Medical Image
      Computing and Computer Assisted Intervention. pp. 699--708 (2021)
    
    \bibitem{cotr}
    Xie, Y., Zhang, J., Shen, C., Xia, Y.: Cotr: Efficiently bridging cnn and
      transformer for 3d medical image segmentation. In: International Conference
      on Medical Image Computing and Computer Assisted Intervention. pp. 171--180.
      Springer (2021)
    
    \bibitem{yangjie}
    Yang, J., Zhang, R., Wang, C., Li, Z., Wan, X., Zhang, L.: Toward unpaired
      multi-modal medical image segmentation via learning structured semantic
      consistency. arXiv preprint arXiv:2206.10571  (2022)
    
    \bibitem{ref9}
    Yang, Y.J., Cho, B.J., Lee, M.J., Kim, J.H., Lim, H., Bang, C.S., Jeong, H.M.,
      Hong, J.T., Baik, G.H.: Automated classification of colorectal neoplasms in
      white-light colonoscopy images via deep learning. Journal of Clinical
      Medicine  \textbf{9}(5), ~1593 (2020)
    
    \bibitem{ref2}
    Zauber, A.G., Winawer, S.J., O'Brien, M.J., Lansdorp-Vogelaar, I., van
      Ballegooijen, M., Hankey, B.F., Shi, W., Bond, J.H., Schapiro, M., Panish,
      J.F., et~al.: Colonoscopic polypectomy and long-term prevention of
      colorectal-cancer deaths. N Engl J Med  \textbf{366},  687--696 (2012)
    
    \bibitem{scan}
    Zhang, R., Li, J., Sun, H., Ge, Y., Luo, P., Wang, X., Lin, L.: Scan:
      Self-and-collaborative attention network for video person re-identification.
      IEEE Transactions on Image Processing  \textbf{28}(10),  4870--4882 (2019)
    
    \bibitem{nnformer}
    Zhou, H.Y., Guo, J., Zhang, Y., Yu, L., Wang, L., Yu, Y.: nnformer: Interleaved
      transformer for volumetric segmentation. arXiv preprint arXiv:2109.03201
      (2021)
    
\end{thebibliography}
\end{document}